# Fast Exploration of Weight Sharing Opportunities for CNN Compression

Etienne Dupuis, Ian O'Connor, Alberto Bosio
*Ecole Centrale de Lyon*
*Institut des Nanotechnologies de Lyon*
Lyon, France
{etienne.dupuis, ian.oconnor, alberto.bosio}@ec-lyon.fr

David Novo
*LIRMM, Université de Montpellier,*
*and CNRS*
Montpellier, France
david.novo@lirmm.fr

*Abstract*—The computational workload involved in Convolutional Neural Networks (CNNs) is typically out of reach for low-power embedded devices. There are a large number of approximation techniques to address this problem. These methods have hyper-parameters that need to be optimized for each CNNs using design space exploration (DSE). The goal of this work is to demonstrate that the DSE phase time can easily explode for state of the art CNN. We thus propose the use of an optimized exploration process to drastically reduce the exploration time without sacrificing the quality of the output.

*Index Terms*—Convolutional Neural Network, Deep Learning, Computer vision, Hardware Accelerator, Design Space Exploration, Approximate Computing, Weight Sharing

## I. INTRODUCTION

Novel computing paradigms and emerging technologies are under investigation to make Convolutional Neural Networks (CNNs) available for low-power devices. Among them, the Approximate Computing paradigm leverages the inherent error resilience of CNNs to improve energy efficiency, by relaxing the need for fully accurate operations. A CNN has a high degree of redundancy in terms of its architecture and parameters, this redundancy is not always necessary for accurate prediction. This observation has paved the way for several highly recognized approximation techniques [1] such as pruning [2], quantization [3], low-rank factorization [4], and weight sharing [2].

In this work, we explore the trade-off between weight sharing (also known as weight clustering) for CNN compression and accuracy loss. In previous work, we have identified that each layer of the network has a different sensitivity to approximation [5], and could benefit from a layer-wise weight sharing.

Thus, we can express weight-sharing as an $N$ variables optimization problem, with $N$ the number of layers of the CNN. Each variable $k_i$ represents the number of shared values of a given layer and is bound to a set of values $k_{range}$, let $k_{tuple} = \{k_i\}$ with $i$ in $[1, N]$. The full exploration complexity is therefore $\mathcal{O}(k_{range}^N)$. Each solution has to be evaluated in terms of *Accuracy Loss* (AL) and *Compression Ratio* (CR). The objective is to maximize CR for a given AL constraint.

For example, given a small CNN like LeNet [6] composed of 5 layers, with a $k_{range} = [1, 256]$ allowing values index to be encoded using up to 8bit. The sequential exhaustive exploration of all $k_{tuple}$ combinations will take $256^5 = \times 10^{12}$ seconds (more than 3000 decades) assuming the clustering (i.e. the selection of shared weights) and the scoring (i.e. CNN AL evaluation) takes 1 second. The evaluation complexity becomes exponentially worse for modern CNNs, which include significantly more layers (e.g., ResNet-152 [7] includes 152 layers). Thus, we need a better method to find the optimal $k_{tuple}$. The research question we are addressing is: *How can we reduce the complexity of the exploration search space for applying a layer-wise weight sharing approximation on a given CNN?*

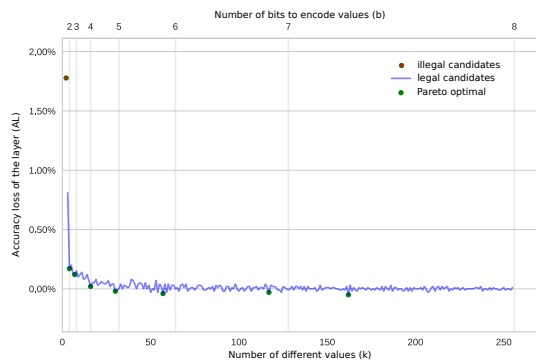

Fig. 1. Studying layer sensitivity to compression by varying the number of shared values

## II. PROPOSED APPROACH

Our approach for optimizing the layer-wise number of shared weights of a weight-shared CNN takes as input a maximum affordable accuracy loss and a trained CNN and outputs a Pareto set of approximated CNN presenting the trade-off between accuracy loss and compression. We can split the optimization problem into two orthogonal sub-problems solved in sequence. The first one, layer optimization, is to select Pareto efficient $k_i$ local to each layer of the network and the second one, network optimization, is to find Pareto efficient combination $k_{tuple}$ of these selected $k_i$.

### A. Layer optimization: Find layer Pareto optimal numbers of unique values

To find the layer Pareto efficient $k_i$, we perform a layer approximation sensitivity analysis by changing the number of shared values and studying the resulting AL and CR. AL is obtained by scoring the CNN using the test set, while CR is obtained as the ratio between the number of shared values over the number of initial values. As we can see in Fig. 1 obtained by varying $k_i$ in the $k_{range} = [1, 256]$ in the third layer of Lenet. The weight sharing AL highly depends on the number of clusters. In order to evaluate CR of the layer, we rely on the number of bit to encode the values $b_{index} = ceil(log_2(k_i))$. This is why we then select Pareto efficient $k_i$ by taking into account the AL and the $b_{index}$. We thus obtain a restrained set of Pareto efficient approximated layer versions $\phi_i$ with $|\phi_i| \leq ceil(log_2(max(k_{range}))$ ($|\phi_i| \leq 8$ in our example). The complexity of the resolution of this first sub-problem is $\mathcal{O}(N * |k_{range}|)$.

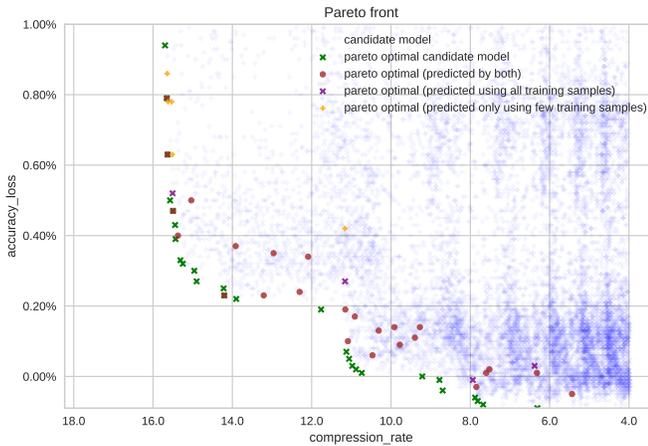

Fig. 2. Comparing Pareto front obtained using CNN evaluation and using prediction model

### B. Network optimization: Find Pareto efficient layer combinations

The second sub-problem is to combine the $k_i$ together in a $k_{tuple}$ to obtain an optimal approximated CNN. The issue is the still large number of possible combinations: $\mathcal{O}(\prod_{i=1}^{N+1} |\phi_i|)$. To drastically reduce the required number of CNN evaluation of the exploration, we use a prediction model giving an analytical approximation of the resulting AL for CNN compressed with a specific $k_{tuple}$, noted $AL_{k_{tuple}}$. The prediction model equation is described by the next formula: $AL_{k_{tuple}} = \sum_{i=1}^{N+1} \alpha_i * AL_{k_i}$, with $AL_{k_i}$ the measured AL during the local optimization for the layer $i$ compressed using $k_i$ shared values, and $\alpha_i$ trained coefficients. This prediction model is trained using standard multivariate linear regression.

### III. PRELIMINARY RESULTS

We started by validating our method on Lenet before going on larger CNN because the reduced number of layers allows for the exhaustive search of all the layer combinations to validate the precision of the prediction model. We used as input a self-trained regular 0.9% top-1 error rate Lenet, and the affordable accuracy loss used is 1%.

By resolving the first sub-problem of layer optimization, we obtained the following layer candidates (only the selected $k_i$ are displayed): layer1 [2, 4, 6, 9, 23, 37], layer2 [3, 7, 14, 26, 47, 78, 130], layer3 [4, 8, 14, 22, 62, 90, 198], layer4 [2, 3, 8, 10, 28, 50, 83] and layer5 [2, 3, 8, 13, 28, 52, 66].

The combination of the different $k_i$ into $k_{tuple}$ represents $14,406$ possible candidate solutions. To validate our method we have performed the scoring of each of them, taking 1 hour on an NVIDIA V100. From the results, it appears that almost 78% of the candidate solution leads to $AL_{k_{tuple}}$ lower than 1%, among these, 27 of them are Pareto optimal, and thus, the most interesting for our optimization problem. To validate that the $AL_{k_{tuple}}$ can be inferred using the prediction model, we fitted it using regression on 80% of the valid candidates for training and the other 20% for testing. We obtained the following set of $\alpha_i$ trained coefficients: $[1.26, 0.78, 0.95, 0.80, 0.92]$. Using this prediction model to evaluate the predicted accuracy of the whole set of candidate result in a very close identified Pareto optimal candidates compared to the original Pareto optimal point as we can see in Fig. 2).

Obtaining $AL_{k_{tuple}}$ of all possible $k_{tuple}$ combinations is not possible for a larger CNN, making this process impossible. Happily, it is not required and we have found that we can train the prediction model on a smaller subset obtained using random subsampling. We have tried using only 0.1% of the dataset (11 samples) for training and the other 99.9% for testing, resulting in a very close Pareto front as we can see Fig. 2). These results prove that we can drastically reduce the necessary number of scoring while keeping a good representation of the candidate behavior in the prediction model, reducing the exploration time by two orders of magnitude.

### IV. CONCLUSION & FUTURE WORK

Our efficient exploration method reduced the number of CNN scoring required to optimize a weight-sharing for a CNN to a few hundred with respect to the initial $10^{12}$. Giving promising results on small CNN, we look forward to applying it to Resnet and Mobilenet, first on CIFAR and then on the Imagenet dataset.

We look forward to using a heuristic metric to avoid the costly scoring step and will take a look at using the inertia (sum of the squared error) of the clustering as a proxy for the accuracy loss.

### V. ACKNOWLEDGEMENT

This work has been funded by the ANR AdequatedDL project (ANR-18-CE23-0012).